\documentclass[10pt,twocolumn,letterpaper]{article}

\usepackage{cvpr}
\usepackage{times}
\usepackage{epsfig}
\usepackage{graphicx}
\usepackage{stmaryrd} 
\usepackage{amsmath,amsthm,amssymb}

\usepackage{myalgorithm}
\usepackage{myalgorithmic}

\theoremstyle{plain}
\newtheorem{lemma}{Lemma}

\theoremstyle{definition}

\usepackage{subfigure}

\usepackage{xspace}
\makeatletter
\DeclareRobustCommand\onedot{\futurelet\@let@token\@onedot}
\def\@onedot{\ifx\@let@token.\else.\null\fi\xspace}
\def\iid{{i.i.d}\onedot}
\def\eg{{e.g}\onedot} 
\def\ie{{i.e}\onedot}

\makeatother

\DeclareMathOperator*{\argmax}{\operatorname{argmax}}

\DeclareMathOperator*{\sign}{\operatorname{sign}}

\newcommand{\Id}{\operatorname{Id}}

\newcommand{\E}{\mathbb{E}}
\newcommand{\F}{\mathcal{F}}

\renewcommand{\H}{\mathcal{H}}
\newcommand{\M}{\mathcal{M}}

\newcommand{\R}{\mathbb{R}}

\newcommand{\X}{\mathcal{X}}
\newcommand{\Y}{\mathcal{Y}}
\newcommand{\Z}{\mathcal{Z}}

\newcommand{\bA}{\mathbf{A}}
\newcommand{\btA}{\mathbf{\tilde A}}
\newcommand{\bAz}{\mathbf{A}}
\newcommand{\btAz}{\mathbf{\tilde A}}

\usepackage[pagebackref=true,breaklinks=true,letterpaper=true,colorlinks,bookmarks=false]{hyperref}

\cvprfinalcopy 

\def\cvprPaperID{1175} 

\usepackage[numbers]{natbib}

\newcommand{\METHODD}{EDD\xspace}
\newcommand{\METHODS}{PredSVM\xspace}

\ifcvprfinal\fi

\renewcommand{\L}{\mathcal{L}}
\begin{document}

\title{Predicting the Future Behavior of a Time-Varying Probability Distribution}

\author{Christoph H. Lampert\\IST Austria\\ {\tt\small chl@ist.ac.at}}

\maketitle


\begin{abstract}
We study the problem of predicting the future, though 
only in the probabilistic sense of estimating a future 
state of a time-varying probability distribution.
This is not only an interesting academic problem, but 
solving this extrapolation problem also has many practical 
application, \eg for training classifiers that have to 
operate under time-varying conditions. 

Our main contribution is a method for predicting the 
next step of the time-varing distribution from a given
sequence of sample sets from earlier time steps. 
For this we rely on two recent machine learning techniques: 
embedding probability distributions into a reproducing kernel 
Hilbert space, and learning operators by vector-valued 
regression. 

We illustrate the working principles and the practical 
usefulness of our method by experiments on synthetic and 
real data. 
We also highlight an exemplary application: training a 
classifier in a domain adaptation setting without having 
access to examples from the test time distribution 
at training time.
\end{abstract}

\vspace*{-\baselineskip}
\section{Introduction}
It is a long lasting dream of humanity to build a machine
that predicts the future. 
For long time intervals, this is likely going to stay 
a dream. 
But for shorter time spans this is not such an 
unreasonable goal.
Humans, in fact, can predict rather reliably how, 
\eg, a video will continue over the next few 
seconds, at least when nothing extraordinary 
happens. 
In this work we aim at making a first step towards giving 
computers similar abilities. 

We study the situation of a time-varying probability distribution 
from which sample sets at different time points are observed.
Our main result is a method for learning an operator that 
captures the dynamics of the time-varying data distribution. 
It relies on two recent techniques: the embedding of probability 
distributions into a reproducing kernel Hilbert space, and 
vector-valued regression. 
By extrapolating the learned dynamics into the future, we obtain 
an estimate of the future distribution. 
This estimate can be used to solve practical tasks, for example, 
learn a classifier that is adapted to the data distribution at 
a future time step, without having access to data from this 
situation already. 
One can also use the estimate to create a new sample set, 
which then can serve as a drop-in replacement for an 
actual sample set from the future. 

\begin{figure*}\centering
\fbox{\includegraphics[width=.99\textwidth]{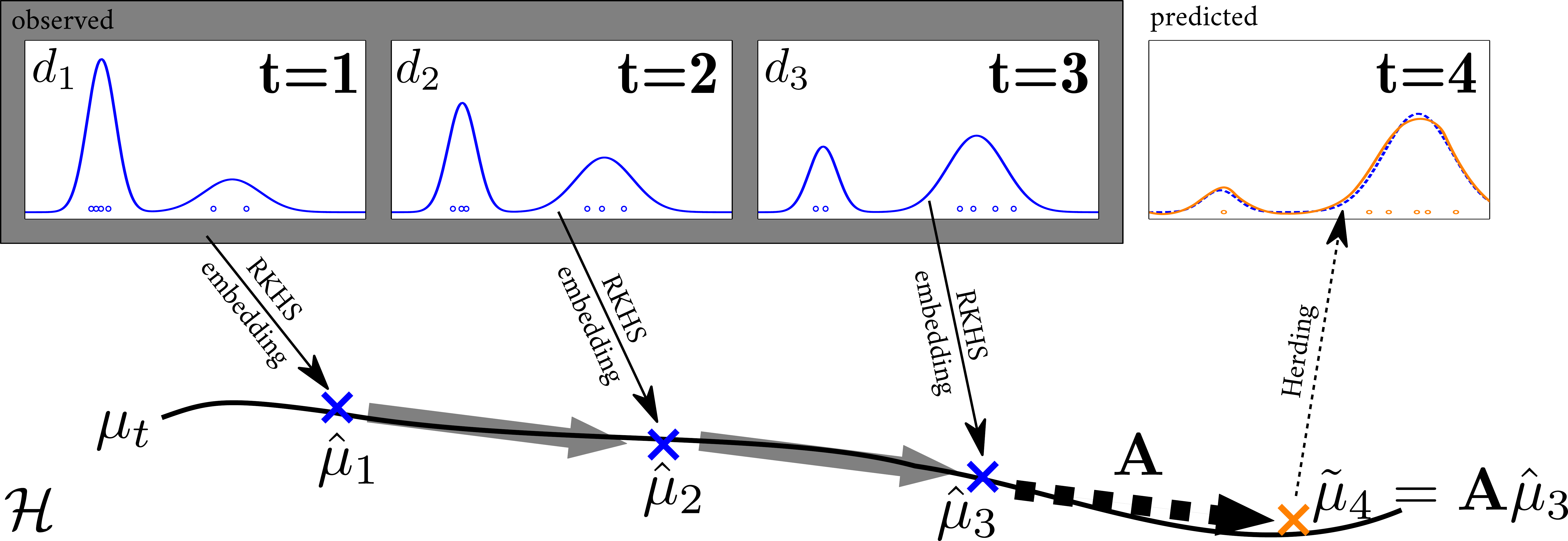}}\smallskip
\caption{Schematic illustration of \METHODD:
we observe samples sets, $S_t$ (blue dots), from a time varying 
probability distribution, $d_t$, at different points of time (blue curves).
Using the framework of RKHS embeddings, we compute their empirical 
kernel mean maps, $\hat\mu_t=\frac{1}{|S_t|}\!\sum_{z\in S_t}\!\phi(z)$ 
in a Hilbert space $\H$.
We learn an operator $\bA:\H\to\H$ that approximates the dynamics from any 
$\hat\mu_t$ to $\hat\mu_{t+1}$ by vector-valued regression (thick gray arrows).
By means of $\bA$ we extrapolate the distribution dynamics beyond 
the last observed distribution (thick dashed arrow), thereby obtaining a 
prediction, $\tilde\mu_{4}$, for the embedding of the unobserved 
\emph{target} distribution $d_{4}$ (dotted blue curve). 
If desired, we apply \emph{herding} (thin dashed arrow) to produce a new 
sample set (orange dots) for the predicted distribution (orange curve).
}\label{fig:schematic}
\end{figure*}

\section{Method}
We first formally define the problem setting of \emph{predicting 
the future of a time-varying probability distribution}. 
Let $\Z$ be a data domain, and let $d_t(z)$ 
for $t\in\mathbb{N}$ be a time-varying data distribution over $z\in\Z$. 
At a fixed point of time, $T$, we assume that we have 
access to sequences of sets, $S_t=\{z^t_1,\dots,z^t_{n_{t}}\}$, 
for $t=1,\dots,T$, that are sampled \iid from the 
respective distributions, $d_1,\dots,d_T$.
Our goal is to construct a distribution, $\tilde d_{T+1}$, 
that is as close as possible to the so far unobserved $d_{T+1}$, 
\ie it provides an estimate of the data distribution one 
step into the future. 
Optionally, we are also interested in obtaining a set, $\tilde S$, 
of samples that are distributed approximated according to 
the unknown $d_{T+1}$. 

Our main contribution is a regression-based method that tackles the 
above problem for the case when the distribution $d_t$ evolves 
smoothly (Sections~\ref{sec:EDD} and~\ref{sec:extension}).
We evaluate this method experimentally in Section~\ref{sec:experiments}.
Subsequently, we show how the ability to extrapolate the distribution 
dynamics can be exploited to improve the accuracy of a classifier 
in a domain adaptation setting without observed data from the 
test time distribution (Section~\ref{sec:PDA}).

\subsection{Extrapolating the Distribution Dynamics}\label{sec:EDD}

We propose a method for \emph{extrapolating the distribution dynamics (EDD)} 
that consist of four steps:
\begin{enumerate}\setlength{\itemsep}{-\parsep}
\item[a)] represent each sample set as a vector in a Hilbert space,
\item[b)] learn an operator that reflects the dynamics between the vectors, 
\item[c)] apply the operator to the last vector in the sequence, 
thereby extrapolating the dynamics by one step,
\item[d)] (optionally) create a new sample set for the extrapolated distribution.
\end{enumerate}
\noindent In the following we discuss the details of each step. 
See Figure~\ref{fig:schematic} for a schematic illustration.

\paragraph{a) RKHS Embedding.}
In order to allow the handling of arbitrary real data, 
we would like to avoid making any domain-specific assumptions, 
such as that the samples correspond to objects in a video, or 
parametric assumptions, such as Gaussianity of the underlying 
distributions.
We achieve this by working in the framework of 
\emph{reproducing kernel Hilbert space (RKHS) embeddings 
of probability distributions}~\cite{smola2007hilbert}.
In this section we provide the most important definitions; 
for a comprehensive introduction see~\cite{song2008learning}.

Let $\mathcal{P}$ denote the set of all probability 
distributions on $\Z$ with respect to some underlying 
$\sigma$-algebra. 
Let $k:\Z\times\Z\to\R$ be a positive definite kernel function with induced RKHS $\H$ 
and feature map $\phi:\Z\to\H$ that fulfills $\|\phi(z)\|\leq 1$ for all $z\in\Z$.
The \emph{kernel mean embedding}, $\mu:\mathcal{P}\to\H$, associated with $k$ 
is defined by 
\begin{align}
p \mapsto \mu(p),\quad \text{ for }\ \mu(p)= \E_{z\sim p(z)} \{\phi(z)\}.
\end{align}

Since we assume $k$ (and therefore $\H$) fixed in this work, 
we also refer to $\mu(p)$ as "the" RKHS embedding of $p$.
We denote by $\M$ the image of $\mathcal{P}$ under $\mu$, 
\ie the set of vectors that correspond to embedded 
probability distributions. 
For \emph{characteristic} kernels, such as the Gaussian, the kernel 
mean map is an \emph{bijection} between $\mathcal{P}$ and $\M$, 
so no information is lost by the embedding operation~\cite{smola2007hilbert}.
In the rest of this section, we will use the term distribution to refer 
to objects either in $\mathcal{P}$ or in $\mathcal{M}$, when it is 
clear from the context which ones we mean.

A useful property of the kernel mean map is 
that it allows us to express the operation of taking 
expected values by an inner product using the 
identity $\E_{p}\{f\} = \langle \mu(p), f\rangle_\H$ 
for any $p\in\mathcal{P}$ and $f\in\H$. 

For a set $S=\{z_1,\dots,z_n\}$ of \iid samples from $p$, 
\begin{align}
S \mapsto \hat\mu(S),\quad \text{for }\ \hat\mu(S) = \frac{1}{n}\sum_{z\in S}\phi(z)
\end{align}
is called the \emph{empirical (kernel mean) embedding} of $S$. 
It is known that under mild conditions on $\H$, the 
empirical embedding, $\hat\mu(S)$, converges with high probability to 
the true embedding, $\mu(p)$, at a rate of $O(1/\sqrt{n})$~\cite{altun2006unifying}. 

\medskip
The first step of \METHODD{} consists of forming the 
embeddings, $\hat\mu_1,\dots,\hat\mu_T$ of the 
observed sample sets, $S_1,\dots,S_T$.
Note that for many interesting kernels the vectors $\hat\mu_t$ 
cannot be computed explicitly, because the kernel feature 
map, $\phi$, is unknown or would require infinite memory 
to be represented. 
However, as we will see later and as it is typical for kernel methods~\cite{lampert-fnt2009}, 
explicit knowledge of the embedding vectors is also not required. 
It is sufficient that we are able to compute their inner products 
with other vectors, and this can be done via evaluations of the 
kernel function.

\paragraph{b) Learning the Dynamics.}
We use \emph{vector-valued regression}~\cite{micchelli2005learning}
to learn a model of the process how the (embedded) distribution 
evolves from one time step to the next.
Vector-valued regression generalizes classical scalar-valued 
regression to the situation, in which the inputs and outputs are 
vectors, \ie the learning of an operator. 
Again, we start by providing a summary of this technique, 
here following the description in~\cite{lian2007nonlinear}.

As basis set in which we search for a suitable operator, we 
define a space, $\F$, of linear operators on $\H$ in the 
following way.
%
Let $\L(\H)$ be the space of all bounded linear operators from $\H$ to $\H$, 
and let $L\!:\!\H\!\times\H\to\L(\H)$ be the \emph{nonnegative $\L(\H)$-valued kernel} 
defined by $L(f,g)=\langle f,g\rangle_{\H}\Id_{\H}$ for any $f,g\in\H$, 
where $\Id_{\H}$ is the identity operator on $\H$. 
Then $L$ can be shown to be the reproducing kernel of an operator-valued RKHS, 
$\F\subseteq\L(\H)$, which contains at least the span of all rank-1 operators, 
$fg^\ast:\H\to\H$, for all $f,g\in\H$, with $g^\ast=\langle g,\cdot\rangle_\H$. 
The inner product between such operators is 
$\langle f_1g_1^\ast,f_2g_2^\ast\rangle_{\F} = \langle f_1,g_1\rangle_\H\langle f_2,g_2\rangle_\H$
for any $f_1,g_1,f_2,g_2\in\H$, and the inner product of all other operators 
in $\F$ can be derived from this by linearity and completeness. 
 
\medskip
As second step of \METHODD we solve a vector-valued regression in order 
to learn a predictive model of the dynamics of the distribution.
For this we assume that the changes of the distributions between
time steps can be approximated by an autoregressive process, \eg $\mu_{t+1} = \bAz\mu_{t} + \epsilon_t$, for some 
operator $\bAz:\H\to\H$, such that the $\epsilon_t$ 
for $t=1,\dots,T$ are independent zero-mean random variables. 
To learn the operator we solve the following least-squares 
functional with regularization constant $\lambda\geq 0$:
\begin{align}
\min_{A\in\F}\,\quad &\sum_{t=1}^{T-1}\  \| \hat\mu_{t+1} - A\hat\mu_{t} \|^2_{\H} \ + \ \lambda\|A\|^2_{\F}.
\label{eq:0th-order}
\intertext{Equation~\eqref{eq:0th-order} has a closed-form solution, }
\btAz&=\sum_{t=1}^{T-1} \hat\mu_{t+1}\sum_{s=1}^{T-1}W_{ts}\hat\mu^\ast_{s},
\label{eq:solution}
\end{align}
with coefficient matrix $W=(K+\lambda I)^{-1}$, where $K\in\R^{(T-1)\times (T-1)}$
is the kernel matrix with entries $K_{st}=\langle \hat\mu_s,\hat\mu_t\rangle_\H$, and $I$ 
is the identity matrix of the same size, see~\cite{lian2007nonlinear} for
the derivation.
Recently, it has been shown that the above regression on distributions 
is consistent under certain technical conditions~\cite{szabo2014twostage}. 
Consequently, if $\bA\in\F$, then the estimated operator, $\btAz$, will 
converge to the true operator, $\bA$, when the number of sample sets and 
the number of samples per set tend to infinity.

%
%

\paragraph{c) Extrapolating the Evolution.}
The third step of \METHODD{} is to extrapolate the dynamics 
of the distribution by one time step. 
With the results of a) and b), all necessary components 
for this are available:
we simply apply the learned operator, $\btA$ to the last observed 
distribution $\hat\mu_T$. 
The result is a prediction, $\tilde\mu_{T+1}=\btA\hat\mu_T$, 
that approximates the unknown target, $\mu_{T+1}$.
From Equation~\eqref{eq:solution} we see that $\tilde\mu_{T+1}$
can be written as a weighted linear combination of the 
observed distributions, 
\begin{align}
&\tilde\mu_{T+1} = \sum_{t=2}^T \beta_{t}\hat\mu_t\ \ \text{with}
\ \ \beta_{t+1}=\sum_{s=1}^{T-1}W_{ts}\langle\hat\mu_s,\hat\mu_T\rangle_{\H},
\label{eq:mu_as_weighted}
\end{align}
for $t=1,\dots,T-1$. 
The coefficients, $\beta_t$, can be computed from the original sample 
sets by means of only kernel evaluations, because 
$\langle\hat\mu_s,\hat\mu_t\rangle_{\H}=\frac{1}{n_sn_t}\sum_{i=1}^{n_s}\sum_{j=1}^{n_t}k(z^s_{i},z^t_{j})$.
The values of $\beta_t$ can be positive or negative, so $\tilde\mu_{T+1}$ is not just 
an interpolation between previous values, but potentially an extrapolation. 
In particular, it can lie outside of the convex hull of the observed distributions.
At the same time, the estimate $\tilde \mu_{T+1}$ is guaranteed to lie in 
the subspace spanned by $\hat\mu_2,\dots,\hat\mu_T$, for which we have 
sample sets available. 
Therefore, so we can compute expected values with respect to $\tilde\mu_{T+1}$ by
forming a suitably weighted linear combinations of the target function at the 
original data points. 
For any $f\in\H$, we have
\begin{align}
\tilde\E_{\tilde\mu_{T+1}}\{f\} &= \langle \tilde\mu_{T+1}, f\rangle_\H \notag
 = 
\sum_{t=2}^T \beta_{t}\big\langle \hat\mu_t, f \big\rangle_{\H} \notag
\\&= 
\sum_{t=2}^T \beta_{t}\frac{1}{n_t}\sum_{i=1}^{n_t}\big\langle \phi(z^t_i), f \big\rangle_{\H} \notag
\\&=
\sum_{t=2}^T \sum_{i=1}^{n_t}\frac{\beta_t}{n_t}f(z^t_i),\label{eq:expectation-future}
\end{align}
where the last identity is due to the fact that $\H$ is the RKHS of 
$k$, which has $\phi$ is its feature map, so $\langle \phi(z), f \rangle_{\H} = f(z)$ for all $z\in\Z$ and $f\in\H$. 
We use the symbol $\tilde\E$ instead of $\E$ to indicate that $\langle \tilde\mu_{T+1}, f\rangle_\H$ 
does not necessarily correspond to the operation of computing an expected value, 
because $\tilde\mu_{T+1}$ might not have a pre-image in the space of probability distributions.
The following lemma shows that $\tilde \mu_{T+1}$ can, nevertheless, act as a reliable proxy for $\mu_{T+1}$:

\begin{lemma}\label{lem:errorbound}
Let $\mu_{T+1}=\bAz\mu_T + \epsilon_T$ and $\tilde\mu_{T+1}=\btAz\hat\mu_{T}$,
for some $\mu_T\in\M$, $\hat\mu_T,\epsilon_T\in\H$ and $\bAz,\btAz\in\F$. 
Then the following inequality holds for all $f\in\H$ with $\|f\|_{\H}\leq 1$,
\begin{align}
|\E_{\mu_{T+1}}\{f\} - \tilde\E_{\tilde\mu_{T+1}}\{f\} |
		&\leq  \ \|\bA\|_{\F}\|\mu_{T}-\hat\mu_{T}\|_{\H} \label{eq:expbound0th} \\
                &\quad   \ + \ \|\bA-\btA\|_{\F}\ + \ \|\epsilon_T\|_{\H}. \nonumber                   
\end{align} 
\end{lemma}
The proof is elementary, using the properties of the inner product and of the RKHS embedding. 

Lemma~\ref{lem:errorbound} quantifies how well $\tilde\mu_{T+1}$ can serve as a drop-in replacement
of $\mu_{T+1}$.
The introduced error will be small, if all three terms on the right hand side 
are small. 
For the first term, we know that this is the case when the number of samples in $S_T$ 
is large enough, since $\|\bA\|_{\F}\|$ is a constant, and we know that the empirical 
distribution, $\hat\mu_T$, converges to the true distribution, $\mu_T$.
Similarly, the second terms becomes small in the limit of many samples set and 
many samples per set, because we know that the estimated operator, $\btA$, 
converges to the operator of the true dynamics, $\bA$, in this case.
Consequently, \METHODD will provide a good estimate of the next distribution time step,
given enough data and if our assumptions about the distribution evolution are 
fulfilled (\ie $\|\epsilon_t\|$ is small). 

\paragraph{d) Generating a Sample Set by Herding.}
Equation~\eqref{eq:expectation-future} suggests a way 
for associating a set of weighted samples with $\tilde\mu_{T+1}$:
\begin{align}
\tilde S_{T+1} = \bigcup_{t=2}^T \Big\{ \frac{\beta_t}{n_t}\cdot z^t_1,\ \dots,\ \frac{\beta_t}{n_t}\cdot z^t_{n_t} \Big\},
\label{eq:weightedsampleset}
\end{align}
where $a\cdot b$ indicates not multiplication but that 
the sample $b$ appears with a weight $a$.
As we will show in Section~\ref{sec:PDA}, this representation is sufficient for 
many purposes, in particular for learning a maximum-margin classifier. 
%
In other situations, however, one might prefer a representation of $\tilde\mu_{T+1}$ 
by uniformly weighted samples, \ie a set $\bar S_{T+1}=\{\bar z_1,\dots,\bar z_{m}\}$ 
such that $\tilde\mu_{T+1}\approx \mu(\bar S_{T+1})= \frac{1}{m}\sum_{i=1}^{m} \phi(\bar z_i)$.
To obtain such a set we propose using the RKHS variant of \emph{herding}~\cite{chen-uai2010}, 
a deterministic procedure for approximating a probability distribution by a set 
of samples.
For any embedded distribution, $\eta\in\M$, herding constructs a sequence 
of samples, $\bar z_1,\bar z_2,\dots,$ by the following rules, 
\begin{align}
\bar z_{1} &= \argmax_{z\in\Z}\big\langle \phi(z), \eta\big\rangle_{\H},
\label{eq:herding}
\\
\bar z_{n} &= \argmax_{z\in\Z}\big\langle \phi(z), \eta\!-\!\frac{1}{n}\sum_{i=1}^{n-1}\phi(\bar z_i)\big\rangle_{\H},
\ \text{for $n\geq 2$.} \nonumber
\end{align}
%
Herding can be understood as an iterative greedy optimization procedure 
for finding examples $\bar z_1,\dots,\bar z_n$ that minimize $\|\eta-\frac{1}{n}\sum_{i=1}^n\phi(\bar z_i)\|_{\H}$~\cite{ICML2012Bach_683}. 
This interpretation shows that the target vector, $\eta$, is not restricted 
to be an embedded distribution, so herding can be applied to arbitrary vectors 
in $\H$.
Doing so for $\tilde\mu_{T+1}$ yields a set $\bar S_{T+1}=\{\bar z_{1},\dots,\bar z_{n_{T+1}}\}$ 
that can act as a drop-in replacement for an actual training set $S_{T+1}$. 
However, it depends on the concrete task whether it is possible 
to compute $\bar S_{T+1}$ in practice, because it requires solving 
multiple pre-image problems~\eqref{eq:herding}, which is not 
always computationally tractable.

A second interesting aspect of herding is that for any $\eta\in\H$, 
the herding approximation always has a pre-image in $\mathcal{P}$
(the empirical distribution defined by $\bar z_1,\dots,\bar z_n$).
Therefore, herding can also be interpreted as an approximate projection 
from $\H$ to $\M$.

\medskip
In Algorithm~\ref{alg:EDD} we provide pseudo-code for \METHODD{}. 
It also shows that despite its mathematical derivation, 
the actual algorithms is easy to implement and execute.

\begin{algorithm}[t]\small
\begin{algorithmic}
\INPUT kernel function $k:\Z\times\Z\to\R$
\INPUT sets $S_t=\{z^t_1,\dots,z^t_{n_t}\}\subset\Z$ for $t=1,\dots,T$
\INPUT regularization parameter $\lambda\geq 0$
\smallskip
\STATE $K\gets$ $(T\!-\!1)\!\times\!(T\!-\!1)$-matrix with entries 

~~~~~~$K_{st} = \frac{1}{n_sn_t}\sum_{i=1}^{n_s}\sum_{j=1}^{n_t} k(z^s_i,z^t_j)$

\smallskip
\STATE $\kappa\gets$ $(T\!-\!1)$-vector with entries 

~~~~~~$\kappa_t = \frac{1}{n_sn_T}\sum_{i=1}^{n_s}\sum_{j=1}^{n_T} k(z^s_i,z^T_j)$

\smallskip
\STATE $\beta^\ast \gets (K+\lambda I)^{-1}\kappa \quad\in\R^{T-1}$ 

\smallskip
\OUTPUT weighted sample set

$\tilde S_{T+1} = \bigcup_{t=2}^T \Big\{ \frac{\beta_t}{n_t}\cdot z^t_1,\ \dots,\ \frac{\beta_t}{n_t}\cdot z^t_{n_t} \Big\},
\text{ with } \beta_t=\beta^{\ast}_{t-1}$
\end{algorithmic}

\smallskip\hrule\smallskip

optional Herding step:

\begin{algorithmic}
\INPUT output size $m$
\STATE $\bar z_1 \gets \argmax_{z\in\Z} \ \sum_{t=2}^T\frac{\beta_t}{n_t}\sum_{i=1}^{n_t} k(z,z^t_i)$

\FOR{$n=2,\dots,m$}
\STATE $\bar z_n\!\gets\!\argmax\limits_{z\in\Z}\Big[\sum_{t=2}^T\frac{\beta_t}{n_t}\sum_{i=1}^{n_t} k(z,z^t_i) \!-\! \frac{1}{n}\sum_{i=1}^{n-1} k(z,\bar z_i)\Big]$
\ENDFOR
\OUTPUT sample set $\bar S_{T+1} = \{ \bar z_1,\dots,\bar z_m \}$
\end{algorithmic}\caption{Extrapolating the distribution dynamics}\label{alg:EDD}
\end{algorithm}

\subsection{Extension to Non-Uniform Weights}\label{sec:extension}
Our above description of \METHODD{}, in particular 
Equation~\eqref{eq:0th-order}, treats all given 
samples sets as equally important.
In practice, this might not be desirable, and one might
want to put more emphasis on some terms in the regression
than on others.
This effect can be achieved by introducing a weight, 
$\gamma_t$, for each of the summands of the least-squares 
problems~\eqref{eq:0th-order}.
Typical choices are $\gamma_t=\rho^{-t}$, for a constant 
$0<\rho< 1$, which expresses a belief that more recent observations
are more trustworthy than earlier ones, or $\gamma_t=\sqrt{n_t}$, 
which encodes that the mean embedding of a sample set is more 
reliable if the set contains more samples.

As in ordinary least squares regression, per-term weights impact 
the coefficient matrix $W$, and thereby the concrete expressions 
for $\beta_t$. However, they do not change the overall structure 
of $\tilde \mu_{T+1}$ as a weighted combination of the observed 
data, so herding and \METHODS{} (see Section~\ref{sec:PDA}) 
training remain possible without structural modifications.

\begin{figure*}[t]\centering
\includegraphics[width=.98\textwidth]{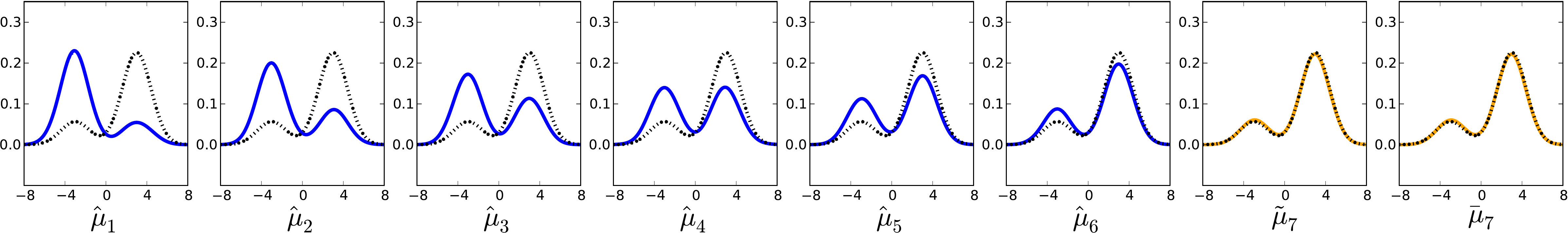}\\\smallskip
\includegraphics[width=.48\textwidth]{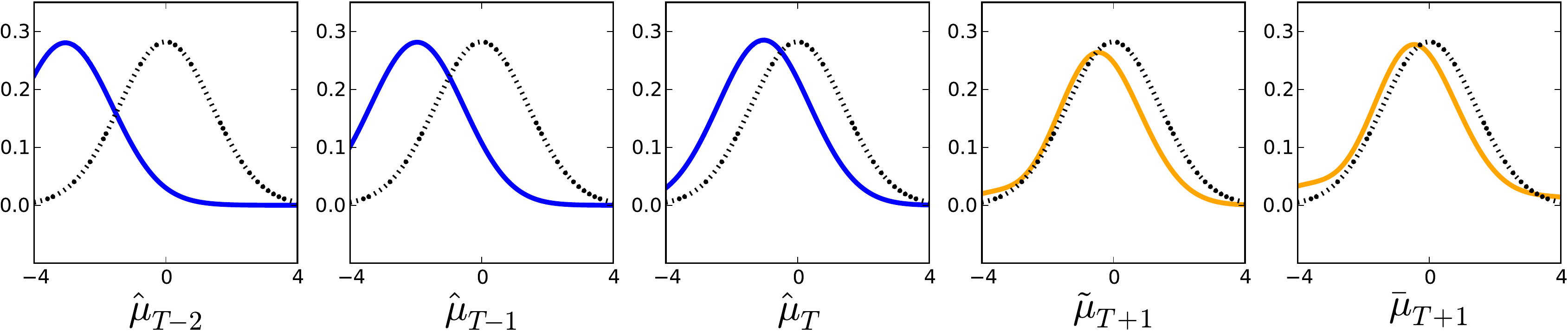}\quad~
\includegraphics[width=.48\textwidth]{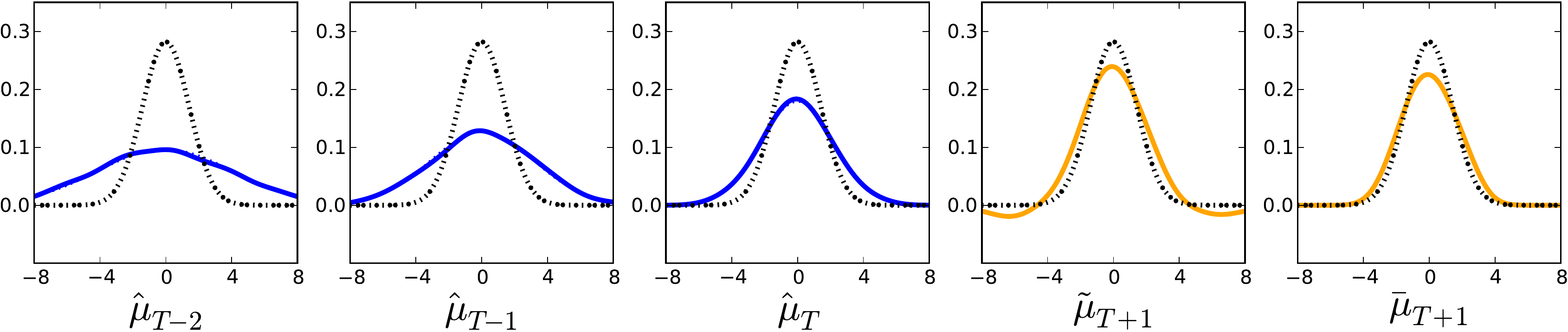}
\caption{Illustration of Experiment 1: mixture of Gaussians with changing proportions (top), 
translating Gaussian (bottom left) and Gaussian with contracting variance (bottom right).
Blue curves illustrate the RKHS embeddings $\hat\mu_{1},\dots,\hat\mu_{T}$ of the given sample sets (no shown). 
The orange curves are \METHODD{}'s prediction of the distribution at time $T+1$, as output 
of the learned operator ($\tilde\mu_T{+1}$) and after additional herding ($\bar\mu_T{+1}$). }
\label{fig:gaussians}
\end{figure*}

\section{Related Work}
To our knowledge, the problem of extrapolating a 
time-varying probability distribution from a set
of samples has not been studied in the literature
before. 
However, a large of body work exists that studies 
related problems or uses related techniques. 

The prediction of future states of a dynamical system or time-variant 
probability distribution is a classical application of probabilistic 
state space models, such as \emph{Kalman filters}~\cite{Kalman60}, and \emph{particle filters}~\cite{gordon1993novel}. 
These techniques aim at modeling the probability of a time-dependent 
system jointly over all time steps. This requires observed data 
in the form of time series, \eg trajectories of moving particles. 
\METHODD{}, on the other hand, learns only the transitions between the 
marginal distribution at one point of time to the marginal distribution 
at the next point of time. 
For this, independent sample sets from different time points
are sufficient. 
The difference between both approaches become apparent, \eg, by looking 
at a system of homogeneously distributed particles that rotate around 
a center. A joint model would learn the circular orbits, while \METHODD{} 
would learn the identity map, since the data distributions are the same 
at any time. 

In the literature of \emph{RKHS embeddings}, a line of work related to
\METHODD{} is the learning of \emph{conditional distributions} by means 
of covariance operators, which has also be interpreted as a vector-valued 
regression task~\cite{lever2012conditional}. 
Given a current distribution and such a conditional model, one could infer 
the marginal distribution of the next time step~\cite{song2009hilbert}. 
Again, the difference to \METHODD{} lies in the nature of the modeled 
distribution and the training data required for this. 
To learn conditional distributions, the training data must consist of pairs 
of data points at two subsequent time points (essentially a minimal trajectory), 
while in the scenario we consider correspondences between samples at 
different time points are not available and often would not even make sense. 
For example, in Section~\ref{sec:experiments} we apply \METHODD{} to 
images of car models from different decades. Correspondence between 
the actual cars depicted in such images do not exist. 

A different line of work aims at predicting the future motion of specific 
objects, such as people or cars, from videos~\cite{kitani2012activity,walker2014patch,yamaguchi2011you,ziebart2009planning}.
These are model-based approaches that target specifically the situation 
of learning trajectories of objects in videos. As such, they can make 
precise predictions about possible locations of objects at future times. 
They are not applicable to the generic situation we are interested, 
however, in which the given data are separate sample sets and the
goal is to predict the future behavior of the underlying probability 
distribution, not of individual objects.

\section{Experiments}\label{sec:experiments}
We report on experiments on synthetic and real data in
order to highlight the working methodology of \METHODD{},
and to show that extrapolating the distribution dynamics 
is possible for real data and useful for practical tasks.

\paragraph{Experiment 1: Synthetic Data.}
First, we perform experiments on synthetic data for which
we know the true data distribution and dynamics, in order
to highlight the working methodology of \METHODD{}.
In each case, we use sample sets of size $n=\{10,100,1000\}$ 
and we use a regularization constant of $\lambda=\frac{1}{n}$. 
Where possible we additionally analytically look at the limit 
case $n\to\infty$, \ie $\hat\mu_t=\mu_t$.
For the RKHS embedding we use a Gaussian kernel with unit variance.

First, we set $d_t = \alpha_t\mathcal{N}(3;1)+(1-\alpha)\mathcal{N}(-3;1)$, 
a mixture of Gaussians distribution with mixture coefficients 
that vary over time as $\alpha_t \in\{ 0.2,0.3,\dots,0.8 \}$. 
Figure~\ref{fig:gaussians} (top) illustrates the results: trained on the first 
six samples sets (blue lines), the prediction by \METHODD{} (orange) match 
almost perfectly the seventh (dashed), with or without herding. 
In order to interpret this result, we first observe that due the 
form of the distributions $d_t$ it is not surprising that $\mu_{T+1}$ 
could be expressed as linear combination of the $\mu_1,\dots,\mu_T$,
provided we allow for negative coefficients. 
What the result shows, however, is that \METHODD{} is indeed able 
to find the right coefficients from the sample sets, indicating that 
the use of an autoregressive model is justified in this case.

Predicting the next time step of a distribution can be expected to be 
harder if not only the values of the density change between time steps 
but also the support. 
We test this by setting $d_t = \mathcal{N}(T\!+\!1\!-\!t;1)$, 
\ie a Gaussian with shifting location of the mean, which we
call the \emph{translation} setting. 
Figure~\ref{fig:gaussians} (bottom left) illustrates the 
last three steps of the total nine observed steps of the 
dynamics (blue) and its extrapolation (orange) with 
and without herding.
One can see that \METHODD{} indeed is able to extrapolate 
the distribution to a new region of the input space:
the mode of the predicted $\tilde\mu_{T+1}$ and $\bar\mu_{T+1}$ 
lie right of the mode of all inputs.
However, the prediction quality is not as good as in the 
mixture setting, indicating that this is in fact a harder 
task. 

Finally, we study a situation where it is not clear on
first sight whether the underlying dynamics has a linear
model: a sequence of Gaussians with decreasing variances, 
$d_t = \mathcal{N}(0 ; T-t+1 )$, which we call 
the \emph{concentration} setting.
The last three steps of the nine observed steps of the dynamics 
and its extrapolation with and without herding are illustrated 
in Figure~\ref{fig:gaussians} (bottom right) in blue and orange,
respectively. 
One can see that despite the likely nonlinearity, \METHODD{} 
is able to predict a distribution that is more concentrated 
(has lower variance) than any of the inputs. 
In this case, we also observe that the predicted distribution 
function has negative values, and that Herding removes those. 

As a quantitative evaluation we report in Tables~\ref{tab:resultsHS} 
and~\ref{tab:resultsKL} how well the predicted distributions correspond 
to the ground truth ones as measured by the Hilbert space (HS) distance 
and the Kullback-Leibler divergences, respectively.
The latter is only possible for \METHODD{} after herding,
when the prediction is a proper probability distribution (non-negative and normalized). 
Besides \METHODD{}, we include the baseline of reusing the last 
observed sample set as a proxy for the next one. To quantify 
how much of the observed distance is due to the prediction step 
and how much is due to an unavoidable sampling error, we also 
report the values for a sample set $S_{T+1}$ of the same size from the true distribution $d_{T+1}$. 

The results confirm that, given sufficiently many samples of the 
earlier tasks,  \METHODD{} is indeed able to successfully 
predict the dynamics of the distribution. 
The predicted distribution $\tilde\mu_{T+1}$ is closer to 
the true distribution $\mu_{T+1}$ than the most similar 
observed distribution, $\hat\mu_T$.
For \emph{translation} and \emph{concentration}, the analytic 
results show that even for $n\to\infty$ the difference is 
non-zero, suggesting that the true dynamics are not exactly 
linear in the RKHS. However, the residual is small compared 
to the measured quantities. 

\begin{table}\setlength{\tabcolsep}{3pt}
\centering
\subtable[\emph{Mixture} setting]{
\begin{tabular}{c|c|c|c|c}
$n$ & \textit{true dist.}& last obs.& \METHODD{} & \METHODD{}+H 
\\\hline
10  & $\mathit{0.13\!\pm\!0.03}$ & $0.13\!\pm\!0.03$ & $0.17\!\pm\!0.02$ & $0.18\!\pm\!0.03$
\\
100 & $\mathit{0.03\!\pm\!0.01}$ & $0.07\!\pm\!0.01$ & $0.05\!\pm\!0.02$ & $0.05\!\pm\!0.02$
\\
1000  & $\mathit{< 0.01}$ & $0.06\!\pm\!0.00$ & $\leq 0.01$ & $< 0.01$
\\
$\infty$ & $\mathit{0.00}$ & $0.07$ & $0.00$ & ---
\end{tabular}}
\subtable[\emph{Translation} setting]{
\begin{tabular}{c|c|c|c|c}
$n$ & \textit{true dist.}& last obs.& \METHODD{} & \METHODD{}+H 
\\\hline
10& $\mathit{0.13\!\pm\!0.12}$&	$0.28\!\pm\!0.21$ &	$0.31\!\pm\!0.17$  &	$0.27\!\pm\!0.18$ 
\\X
100  &	$\mathit{0.04\!\pm\!0.04}$&	$0.27\!\pm\!0.12$ &	$0.20\!\pm\!0.10$ &	$0.18\!\pm\!0.11$ 
\\
1000  &	$\mathit{0.01\!\pm\!0.01}$ &	$0.26\!\pm\!0.07$  &	$0.14\!\pm\!0.06$ &	$0.13\!\pm\!0.06$ 
\\
$\infty$ & $\mathit{0.00}$ & 0.27  & 0.09 &---
\end{tabular}}
\subtable[\emph{Concentration} setting]{
\begin{tabular}{c|c|c|c|c}
$n$ & \textit{true dist.}& last obs.& \METHODD{} & \METHODD{}+H 
\\\hline
10&		$\mathit{0.13\!\pm\!0.12}$ &	$0.25\!\pm\!0.20$ &	$0.32\!\pm\!0.17$ &	$0.32\!\pm\!0.18$
\\
100  &	$\mathit{0.04\!\pm\!0.04}$ &	$0.20\!\pm\!0.10$ &	$0.22\!\pm\!0.12$ &$0.22\!\pm\!0.12$ 
\\
1000  &		$\mathit{0.01\!\pm\!0.01}$ &$0.19\!\pm\!0.06$ &	$0.15\!\pm\!0.07$ &$0.15\!\pm\!0.07$ 
\\
$\infty$ & $\mathit{0.00}$ & $0.19$ & 0.07 & --- 
\end{tabular}}
\caption{Approximation quality of \METHODD{}, \METHODD{} with herding (\METHODD+H) and baselines measured in RKHS norm (lower values are better)
for different synthetic settings. For details, see Section~\ref{sec:experiments}.}\label{tab:resultsHS}
\end{table}

\begin{table}
\centering
\subtable[\emph{Mixture} setting]{
\begin{tabular}{c|c|c|c}
$n$ &  \textit{true dist.}& last obs.&   \METHODD{}+H
\\\hline
10 &	$\mathit{0.08 \pm 0.05}$ &	$0.08 \pm 0.03$  &	$0.17 \pm 0.05$
\\
100 &	$\mathit{<0.01}$ &	$0.03 \pm 0.01$ & $0.02 \pm 0.01$
\\
1000 & $\mathit{<0.001}$ &	$0.02 \pm 0.00$ & $<0.005$
\end{tabular}}
\subtable[\emph{Translation} setting]{
\begin{tabular}{c|c|c|c}
$n$ &  \textit{true dist.}& last obs.&   \METHODD{}+H
\\\hline
10 &	$\mathit{0.05 \pm 0.05}$ &	$0.29 \pm 0.19$  &	$0.28 \pm 0.10$
\\
100 &	$\mathit{<0.005}$ &	$0.26 \pm 0.05$ & $0.11 \pm 0.04$ 
\\
1000 & $\mathit{<0.001}$ &	$0.25 \pm 0.02$ & $0.07 \pm 0.02$ 
\end{tabular}}
\subtable[\emph{Concentration} setting]{
\begin{tabular}{c|c|c|c}
$n$ & \textit{true dist.}& last obs.&  \METHODD{}+H
\\\hline
10 &	$\mathit{0.05 \pm 0.05}$ &	$0.23 \pm 0.13$ &	$0.56 \pm 0.18$
\\
100 &	$\mathit{<0.005}$  &	$0.16 \pm 0.04$ &	$0.20 \pm 0.06$ 
\\
1000 &	$\mathit{<0.001}$ & $0.16 \pm 0.01$ &$0.06 \pm 0.02$ 
\end{tabular}}
\caption{Approximation quality of \METHODD{}, \METHODD{} with herding (\METHODD+H) and baselines measured by KL divergence (lower values are better)
for different synthetic settings. For details, see Section~\ref{sec:experiments}.}\label{tab:resultsKL}
\end{table}


\begin{table*}[t]
\small\centering\setlength{\tabcolsep}{4pt}
\subfigure[Histogram intersection kernel]{\begin{tabular}{l|c|c|c}
\textbf{YouTube} & EDD & last seg. & all seg. 
\\\hline
birthday(151) 
& $\mathbf{0.15 \pm 0.005}$  & $         0.16 \pm 0.006 $  & $         0.16 \pm 0.005 $  \\\hline
parade(119) 
& $\mathbf{0.13 \pm 0.006}$  & $         0.15 \pm 0.008 $  & $         0.15 \pm 0.007 $  \\\hline
picnic(85) 
& $\mathbf{0.13 \pm 0.007}$  & $         0.15 \pm 0.009 $  & $         0.15 \pm 0.008 $  \\\hline
show(200) 
& $\mathbf{0.14 \pm 0.004}$  & $         0.15 \pm 0.005 $  & $         0.16 \pm 0.005 $  \\\hline
sports(258) 
& $\mathbf{0.14 \pm 0.004}$  & $         0.15 \pm 0.004 $  & $         0.16 \pm 0.004 $  \\\hline
wedding(90) 
& $\mathbf{0.16 \pm 0.007}$  & $         0.17 \pm 0.009 $  & $         0.18 \pm 0.007 $  \smallskip
\\
\textbf{Kodak} & EDD & last seg. & all seg. 
\\\hline
birthday(16) 
& $\mathbf{0.17 \pm 0.015}$  & $         0.20 \pm 0.022 $  & $         0.18 \pm 0.014 $  \\\hline
parade(14) 
& $         0.15 \pm 0.023 $  & $         0.17 \pm 0.030 $  & $         0.17 \pm 0.022 $  \\\hline
picnic(6) 
& $         0.17 \pm 0.028 $  & $         0.17 \pm 0.031 $  & $         0.20 \pm 0.027 $  \\\hline
show(55) 
& $\mathbf{0.20 \pm 0.011}$  & $         0.23 \pm 0.013 $  & $         0.21 \pm 0.011 $  \\\hline
sports(74) 
& $         0.15 \pm 0.006 $  & $         0.16 \pm 0.007 $  & $         0.16 \pm 0.007 $  \\\hline
wedding(27) 
& $\mathbf{0.18 \pm 0.011}$  & $         0.21 \pm 0.013 $  & $         0.19 \pm 0.011 $  \smallskip
\\
\end{tabular}}
\qquad
\subfigure[RBF-$\chi^2$ kernel]{\begin{tabular}{l|c|c|c}
\textbf{YouTube} & EDD & last seg. & all seg. 
\\\hline
birthday(151) 
& $\mathbf{0.17 \pm 0.006}$  & $         0.19 \pm 0.007 $  & $         0.19 \pm 0.006 $  \\\hline
parade(119) 
& $\mathbf{0.15 \pm 0.007}$  & $         0.17 \pm 0.009 $  & $         0.18 \pm 0.008 $  \\\hline
picnic(85) 
& $\mathbf{0.15 \pm 0.008}$  & $         0.17 \pm 0.010 $  & $         0.18 \pm 0.010 $  \\\hline
show(200) 
& $\mathbf{0.17 \pm 0.005}$  & $         0.18 \pm 0.005 $  & $         0.20 \pm 0.006 $  \\\hline
sports(258) 
& $\mathbf{0.16 \pm 0.004}$  & $         0.17 \pm 0.005 $  & $         0.20 \pm 0.005 $  \\\hline
wedding(90) 
& $\mathbf{0.18 \pm 0.008}$  & $         0.20 \pm 0.010 $  & $         0.21 \pm 0.008 $  \smallskip
\\
\textbf{Kodak} & EDD & last seg. & all seg. 
\\\hline
birthday(16) 
& $         0.20 \pm 0.018 $  & $         0.24 \pm 0.027 $  & $         0.22 \pm 0.015 $  \\\hline
parade(14) 
& $         0.18 \pm 0.027 $  & $         0.20 \pm 0.035 $  & $         0.20 \pm 0.025 $  \\\hline
picnic(6) 
& $         0.19 \pm 0.029 $  & $         0.19 \pm 0.032 $  & $         0.24 \pm 0.031 $  \\\hline
show(55) 
& $\mathbf{0.23 \pm 0.012}$  & $         0.26 \pm 0.015 $  & $         0.25 \pm 0.013 $  \\\hline
sports(74) 
& $\mathbf{0.17 \pm 0.007}$  & $         0.18 \pm 0.008 $  & $         0.20 \pm 0.008 $  \\\hline
wedding(27) 
& $         0.22 \pm 0.012 $  & $         0.24 \pm 0.015 $  & $         0.23 \pm 0.013 $  \smallskip
\\
\end{tabular}}
\caption{Experiment 2: Distance between last video segment and its 
prediction by \METHODD, the last observed segment (last seg.) and 
the union of all segments (all seg.).}
\label{tab:video}
\end{table*}

\paragraph{Experiment 2: Real World Data.}
In a second set of experiments, we test \METHODD's suitability 
for real data by applying to video sequences from~\cite{duan2012visual}. 
The dataset consists of 1121 video sequences of six 
semantic categories, \emph{birthday, parade, picnic, show, sports}, and \emph{wedding}, 
from two sources, \emph{Kodak} and \emph{YouTube}.
Each video is represented by a collection of spatio-temporal 
interest points (STIPs) with 162-dimensional feature 
vectors.\footnote{\url{http://vc.sce.ntu.edu.sg/index_files/VisualEventRecognition/features.html}}

For each video, except six that were less than one second 
long, we split the STIPs into groups by creating segments 
of 10 frames each. 
Different segments have different numbers of samples, 
because the STIPs were detected based on the response 
of an interest operator. 
Different video also show a strong diversity in this 
characteristics: the number of per STIPs per segment 
varies between 1 and 550, and the number of segments 
per video varies between 3 and 837.

As experimental setup, we use all segments of a movie except 
the last one as input sets for \METHODD{}, and we measure 
the distance between the predicted next distribution and 
the actual last segment. 
Table~\ref{tab:video} shows the results split by data source
and category for two choices of kernels: the \emph{RBF-$\chi^2$ kernel}, $k(z,\bar z) = \exp( -\frac12\chi^2(z,\bar z))$ 
for $\chi^2(z,\bar z)=\frac{1}{d}\sum_{i=1}^{d} \frac{(z_i-\bar z_i)^2}{\frac12(z_i+\bar z_i)}$,
and the \emph{histogram intersection kernel}, $k(z,\bar z) = \frac{1}{d}\sum_{i=1}^d \min(z_i,\bar z_i)$,
both for $z,\bar z\in\R^{d}_{+}$.
For each data source and category we report the average and standard 
error of the Hilbert-space distance between distribution.
As baselines, we compare against re-using the last observed 
segment, \ie not extrapolating, and against the distribution 
obtained from merging all segments, \ie the global video 
distribution. 
One can see that the predictions by \METHODD{} are closer 
to the true evolution of the videos than both baselines 
in all cases but two, in which it is tied with using the 
last observation. 
The improvement is statistically significant (bold print) 
to a 0.05 level according to Wilcoxon signed rank test
with multi-test correction, except for some cases with 
only few sequences. 

\section{Application: Predictive Domain Adaptation}\label{sec:PDA}
We are convinced that being able to extrapolate a time-varying 
probability distribution into the future will be useful 
for numerous practical applications. 
As an example, we look at one specific problem: learning a 
classifier under distribution drift, when for training the 
classifier data from the time steps $t=1,\dots,T$ is 
available, but by the time the classifier is applied 
to its target data, the distribution has moved on to 
time $t=T\!+\!1$.
A natural choice to tackle this situation would be by 
using \emph{domain adaptation} techniques~\cite{jiang2008literature}.
However, those typically require that at least unlabeled data 
from the target distribution is available, which in practice
might not be the case.
For example, in an online prediction setting, such as spam filtering, 
predictions need to be made on the fly. 
One cannot simply stop, collect data from the new data distribution, 
and retrain the classifiers.
Instead, we show how \METHODD can be used to train a maximum margin 
classifier for data distributed according to $d_{T+1}$, with only
data from $d_{1}$ to $d_{T}$ available.
We call this setup \emph{predictive domain adaptation (PDA)}.

Let $S_t=\{(x^t_1,y^t_1),\dots,(x^t_{n_t},y^t_{n_t})\}$ for $t=1,\dots,T$, 
be a sequence of labeled training sets, where $\X$ is the input space, \eg 
images, and $\Y=\{1,\dots,K\}$ is the set of class labels. 
For any kernel $k_{\X}(x,\bar x)$ on $\X$, we form a joint 
kernel, $k((x,y),(\bar x,\bar y))=k_{\X}(x,\bar x)\llbracket y=\bar y\rrbracket$
on $\X\times\Y$ and we apply \METHODD{} for $\Z=\X\times\Y$. 
The result is an estimate of the next time step of the joint 
probability distribution, $d_{T+1}(x,y)$, as a vector, 
$\tilde\mu_{T+1}$, or in form of a weighted sample 
set, $\tilde S_{T+1}$.

To see how this allows us to learn a better adapted classifier, 
we first look at the situation of classification with 0/1-loss
function, $\ell(y,\bar y)=\llbracket y\neq \bar y\rrbracket$. 
If a correctly distributed training $S_{T+1}$ of size $n_{T+1}$ were 
available, one would aim for minimizing the regularized risk functional 
\begin{align}
\frac12\|w\|^2 + \frac{C}{n_{T+1}}\!\!\sum_{(x_i,y_i)\in S_{T+1}}\!\!\!\!\!\! \ell(\,y_i,\sign \langle w,\psi(x_i)\rangle\,), \label{eq:regularizedrisk}
\end{align}
where $C$ is a regularization parameter and $\psi$ is any 
feature map, not necessarily the one induced by $k_{\X}$. 
To do so numerically, one would bound the loss by a convex surrogate, 
such as the hinge loss, $\max\{ 0, 1-y\langle w,\psi(x)\rangle\}$, 
which make the overall optimization problem convex and therefore 
efficiently solvable.

In the PDA situation, we do not have a training set $S_{T+1}$, but 
we do have a prediction $\tilde S_{T+1}$ provided by \METHODD 
in the form of Equation~\eqref{eq:weightedsampleset}. 
%
%
Therefore, instead of the empirical average in~\eqref{eq:regularizedrisk},
we can form a predicted empirical average using the weighted samples in $\tilde S_{T+1}$.
This leads to the \emph{predicted regularized risk functional}, 
\begin{align}
\frac12\|w\|^2 + C\sum_{t=2}^T \frac{\beta_t}{n_t} \sum_{i=1}^{n_t} \ell(\,y^t_i,\sign \langle w,\psi(x^t_i)\rangle\,),
\label{eq:predictedregularizedrisk}
\end{align}
that we would like to minimize. 

\begin{figure*}\centering
\begin{tabular}{ccc}
BMW & Mercedes & VW \\
\includegraphics[width=.29\textwidth]{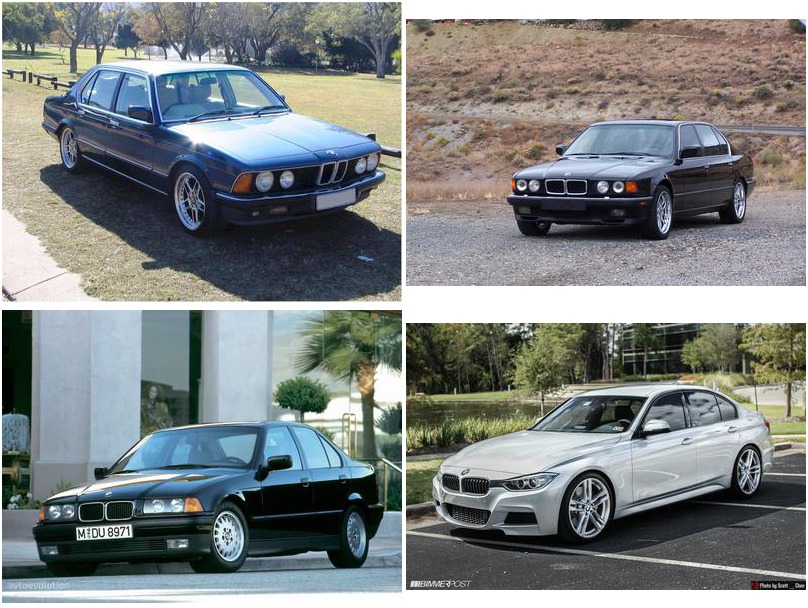}\quad&
\includegraphics[width=.29\textwidth]{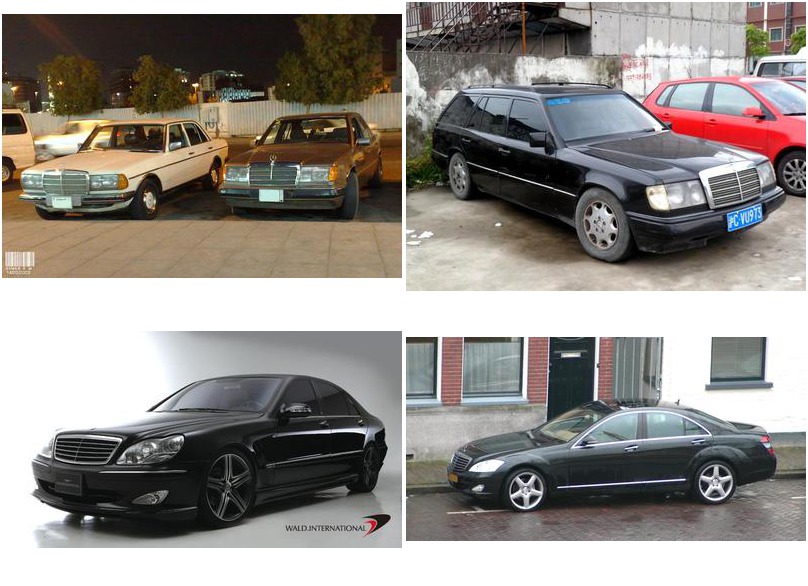}\quad&
\includegraphics[width=.29\textwidth]{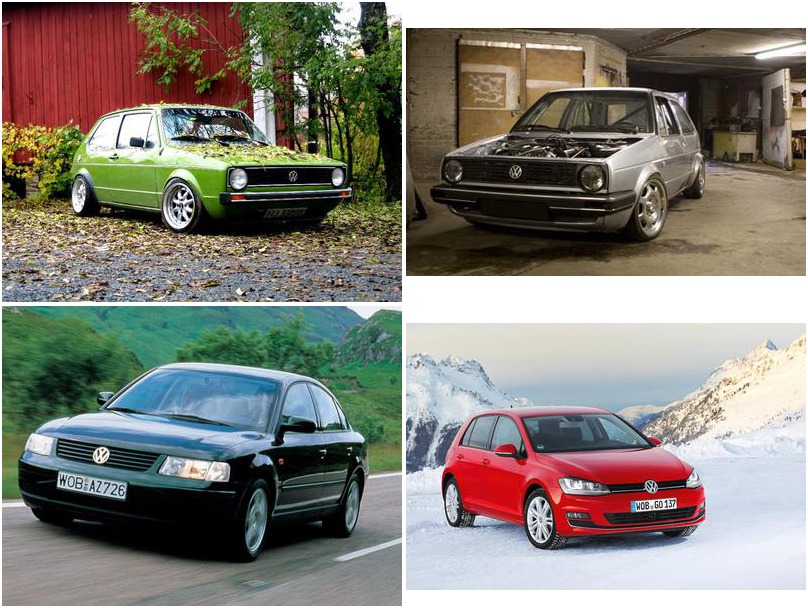}
\end{tabular}
\caption{Example images from \emph{CarEvolution} dataset~\cite{RematasICCVWS13}. 
The goal is to classify images by their manufacturer (BMW, Mercedes, VW). 
Each block shows one image from each the four groups: \emph{1970s} 
(top left), \emph{1980s} (top right), \emph{1990s} (bottom left), and \emph{later} 
(bottom right).}
\end{figure*}

In contrast to the expression~\eqref{eq:regularizedrisk}, 
replacing the 0/1-loss by the hinge loss does not lead to 
a convex upper bound of~\eqref{eq:predictedregularizedrisk}, 
because the coefficients $\beta_t$ can be either positive or negative.
However, we can use that $\ell(\, y, \bar y\,) = 1-\ell(\, -y, \bar y\,)$,
and obtain an equivalent expression for the loss term with only 
positive weights, $\sum_{t=2}^T b_{t} \sum_{i=1}^{n_t} \ell(\, \bar y_{ti},\sign f(x_{ti})\,) + c$, 
where $b_{t}=|\beta_{t}|/n_t$ and $\bar y_{ti}=(\sign\beta_{t})y_{ti}$. 
The constant $c=\sum_t\beta_t$ plays no role for the optimization procedure, 
so we drop it from the notation for the rest of this section.
Now bounding each 0/1-loss term by the corresponding Hinge 
loss yields a convex upper bound of the predicted risk,
\begin{align}
\frac{1}{2}\|w\|^2 +C\sum_{t=2}^T b_t \sum_{i=1}^{n_t} \max\{ 0, 1\!-\!\bar y_{ti}\langle w,\psi(x^t_i)\rangle\}.
\label{eq:FPDA-primalSVM}
\end{align}

Minimizing it corresponds to training a support vector machine 
with respect to the predicted data distribution, which we refer
to as \METHODS.
It can be done by standard SVM packages that support per-sample 
weight, such as libSVM~\cite{libsvm}. 

\paragraph{Experiment 3: Predictive Domain Adaptation.}
To demonstrate the usefulness of training a classifier on a predicted 
data distribution, we perform experiments on the 
\emph{CarEvolution}~\cite{RematasICCVWS13} data set.\footnote{\url{http://homes.esat.kuleuven.be/~krematas/VisDA/CarEvolution.html}} 
It consists of 1086 images of cars, each annotated by the 
car manufacturer (\textit{BMW}, \textit{Mercedes} or \textit{VW}) and 
the year in which the car model was introduced (between 1972 and 2013). 

The data comes split into source data (years 1972--1999) and target data (years 2000-2013). 
We split the \emph{source} part further into three decades: 1970s, 1980s, 1990s. 
Given these groups, our goal is to learn a linear \METHODS to distinguish between the manufacturers in 
the \emph{target} part.
We also perform a second set of experiments, where we split the target set 
further into models from the 2000s and models from the 2010s, and we learn 
a linear \METHODS with the 1970s as target and the other tasks in inverse 
order as sources.
As baseline, we use SVMs that were trained on any of the observed tasks, 
as well as an SVMs trained all the union of all source tasks. 
In all cases we choose the SVM parameter, $C\in\{10^0,\dots,10^{6}\}$, by five-fold 
cross validation on the respective training sets.

Table~\ref{tab:car-results} summarizes the results for two different feature representations: 
Fisher vectors~\cite{perronnin2010improving} and $L^2$-normalized 
DeCAF features~\cite{donahue2014decaf}.
In both cases, \METHODS is able to improve over the baselines. 
Interestingly, the effect is stronger for the DeCAF features, even though 
these were reported to be less affected by visual domain shifts. 
We plan to explore this effect in further work.
%


\begin{table}\centering
\begin{tabular}{c|c|c}
method                & FVs & decaf 
\\\hline
1970s\,$\to$\,$\geq$2000s & 39.3\% & 38.2\% 
\\
1980s\,$\to$\,$\geq$2000s & 43.8\% & 48.4\% 
\\
1990s\,$\to$\,$\geq$2000s & 49.0\% & 52.4\% 
\\
  all\,$\to$\,$\geq$2000s & 51.2\% & 52.1\% 
\\
\METHODS{} (temporal order) & 51.5\% & 56.2\% 
\smallskip\\
method                & FVs & decaf 
\\\hline
2010s\,$\to$\,1970s & 33.5\% & 34.0\% 
\\
2000s\,$\to$\,1970s & 31.6\% & 42.7\%
\\
1990s\,$\to$\,1970s & 46.1\% & 46.6\%
\\
1980s\,$\to$\,1970s & 44.7\% & 33.5\%
\\
  all\,$\to$\,1970s & 46.1\% & 49.0\%
\\
\METHODS{} (reverse order) & 48.5\% & 54.4\%
\end{tabular}
\caption{Classification accuracy of PDA-SVM and baseline methods on \emph{CarEvolution} data set (higher is better). 
Top: temporal order, bottom: reverse order. See Section~\ref{sec:experiments} for details.}\label{tab:car-results}
\end{table}

\section{Summary and Discussion}\vspace{-.5\baselineskip}
In this work, we have introduced the task of predicting the 
future evolution of a time-varying probability distribution. 
We described a method that, given a sequence of observed 
samples set, extrapolates the distribution dynamics 
by one step. 
Its main components are two recent techniques from machine learning:
the embeddings of probability distributions into a Hilbert space, 
and vector-valued regression. 
Furthermore, we showed how the predicted distribution obtained
from the extrapolation can be used to learn a classifier for a 
data distribution from which no training examples is available, 
not even unlabeled ones.

Our experiments on synthetic and real data gave insight into the
working methodology of \METHODD{} and showed that it is 
--to some extend-- possible to predict the next state of a 
time-varying distribution from sample sets of earlier time steps, 
and that this can be useful for learning better classifiers. 
One shortcoming of our method is that currently it is restricted 
to equally spaced time steps and that the extrapolation is only 
by a single time unit. 
We plan to extend our framework to more flexible situations, 
including distributions with a continuous time parameterization. 

\subsection*{Acknowledgements}
This work was funded in parts by the European Research Council 
under the European Unions Seventh Framework Programme 
(FP7/2007-2013)/ERC grant agreement no 308036.


\begin{thebibliography}{10}\itemsep=-1pt

\bibitem{altun2006unifying}
Y.~Altun and A.~Smola.
\newblock Unifying divergence minimization and statistical inference via convex
  duality.
\newblock In {\em Workshop on Computational Learning Theory (COLT)}, 2006.

\bibitem{ICML2012Bach_683}
F.~Bach, S.~Lacoste-Julien, and G.~Obozinski.
\newblock On the equivalence between herding and conditional gradient
  algorithms.
\newblock In {\em International Conference on Machine Learing (ICML)}, 2012.

\bibitem{libsvm}
C.-C. Chang and C.-J. Lin.
\newblock {LIBSVM}: {A} library for support vector machines.
\newblock {\em ACM Transactions on Intelligent Systems and Technology},
  2:27:1--27:27, 2011.

\bibitem{chen-uai2010}
Y.~Chen, M.~Welling, and A.~J. Smola.
\newblock Super-samples from kernel herding.
\newblock In {\em Uncertainty in Artificial Intelligence (UAI)}, 2010.

\bibitem{donahue2014decaf}
J.~Donahue, Y.~Jia, O.~Vinyals, J.~Hoffman, N.~Zhang, E.~Tzeng, and T.~Darrell.
\newblock {DeCAF}: {A} deep convolutional activation feature for generic visual
  recognition.
\newblock In {\em International Conference on Machine Learing (ICML)}, 2014.

\bibitem{duan2012visual}
L.~Duan, D.~Xu, I.-H. Tsang, and J.~Luo.
\newblock Visual event recognition in videos by learning from web data.
\newblock {\em IEEE Transactions on Pattern Analysis and Machine Intelligence
  (T-PAMI)}, 34(9):1667--1680, 2012.

\bibitem{gordon1993novel}
N.~J. Gordon, D.~J. Salmond, and A.~F.~M. Smith.
\newblock Novel approach to nonlinear/non-{G}aussian {B}ayesian state
  estimation.
\newblock In {\em IEE Proceedings F (Radar and Signal Processing)}, 1993.

\bibitem{jiang2008literature}
J.~Jiang.
\newblock A literature survey on domain adaptation of statistical classifiers.
\newblock \url{http://sifaka.cs.uiuc.edu/jiang4/domain_adaptation/survey},
  2008.

\bibitem{Kalman60}
R.~E. Kalman.
\newblock A new approach to linear filtering and prediction problems.
\newblock {\em Journal of Basic Engineering}, 82:35--45, 1960.

\bibitem{kitani2012activity}
K.~M. Kitani, B.~D. Ziebart, J.~A. Bagnell, and M.~Hebert.
\newblock Activity forecasting.
\newblock In {\em European Conference on Computer Vision (ECCV)}, pages
  201--214, 2012.

\bibitem{lampert-fnt2009}
C.~H. Lampert.
\newblock Kernel methods in computer vision.
\newblock {\em Foundations and Trends in Computer Graphics and Vision},
  4(3):193--285, 2009.

\bibitem{lever2012conditional}
G.~Lever, L.~Baldassarre, S.~Patterson, A.~Gretton, M.~Pontil, and
  S.~Gr{\"u}new{\"a}lder.
\newblock Conditional mean embeddings as regressors.
\newblock In {\em International Conference on Machine Learing (ICML)}, 2012.

\bibitem{lian2007nonlinear}
H.~Lian.
\newblock Nonlinear functional models for functional responses in reproducing
  kernel {H}ilbert spaces.
\newblock {\em Canadian Journal of Statistics}, 35(4):597--606, 2007.

\bibitem{micchelli2005learning}
C.~A. Micchelli and M.~Pontil.
\newblock On learning vector-valued functions.
\newblock {\em Neural Computation}, 17(1):177--204, 2005.

\bibitem{perronnin2010improving}
F.~Perronnin, J.~S{\'a}nchez, and T.~Mensink.
\newblock Improving the {F}isher kernel for large-scale image classification.
\newblock In {\em European Conference on Computer Vision (ECCV)}, 2010.

\bibitem{RematasICCVWS13}
K.~Rematas, B.~Fernando, T.~Tommasi, and T.~Tuytelaars.
\newblock Does evolution cause a domain shift?
\newblock In {\em ICCV Workshop on Visual Domain Adaptation and Dataset Bias},
  2013.

\bibitem{smola2007hilbert}
A.~Smola, A.~Gretton, L.~Song, and B.~Sch{\"o}lkopf.
\newblock A {H}ilbert space embedding for distributions.
\newblock In {\em International Conference on Algorithmic Learning Theory
  (ALT)}, 2007.

\bibitem{song2008learning}
L.~Song.
\newblock {\em Learning via Hilbert space embedding of distributions}.
\newblock PhD thesis, University of Sydney, 2008.

\bibitem{song2009hilbert}
L.~Song, J.~Huang, A.~Smola, and K.~Fukumizu.
\newblock {H}ilbert space embeddings of conditional distributions with
  applications to dynamical systems.
\newblock In {\em International Conference on Machine Learing (ICML)}, 2009.

\bibitem{szabo2014twostage}
Z.~Szabo, A.~Gretton, B.~Poczos, and B.~Sriperumbudur.
\newblock Learning theory for distribution regression.
\newblock arXiv:1411.2066 [math.ST], 2014.

\bibitem{walker2014patch}
J.~Walker, A.~Gupta, and M.~Hebert.
\newblock Patch to the future: Unsupervised visual prediction.
\newblock In {\em Conference on Computer Vision and Pattern Recognition
  (CVPR)}, 2014.

\bibitem{yamaguchi2011you}
K.~Yamaguchi, A.~C. Berg, L.~E. Ortiz, and T.~L. Berg.
\newblock Who are you with and where are you going?
\newblock In {\em Conference on Computer Vision and Pattern Recognition
  (CVPR)}, 2011.

\bibitem{ziebart2009planning}
B.~D. Ziebart, N.~Ratliff, G.~Gallagher, C.~Mertz, K.~Peterson, J.~A. Bagnell,
  M.~Hebert, A.~K. Dey, and S.~Srinivasa.
\newblock Planning-based prediction for pedestrians.
\newblock In {\em IEEE/RSJ International Conference on Intelligent Robots and
  Systems (IROS)}, 2009.

\end{thebibliography}
\end{document}